# Achieving Meaningful Collaboration: Worker-centered Design of a Physical Human-Robot Collaborative Blending Task

Nicky Mol[1], Luka Peternel[1], Alessandro Ianniello[2], Denis Zatyagov[3], Auke Nachenius[3], Stephan Balvert[3], J. Micah Prendergast[1], Sara Muscolo[2], Olger Siebinga[2], Eva Verhoef[3], Deborah Forster[1,3], David A. Abbink[1,2,3]

*Abstract*— The use of robots in industrial settings continues to grow, driven by the need to address complex societal challenges such as labor shortages, aging populations, and ever-increasing production demands. In this abstract, we advocate for (and demonstrate) a transdisciplinary approach when considering robotics in the workplace. Transdisciplinarity emphasizes the integration of academic research with pragmatic expertise and embodied experiential knowledge, that prioritize values such as worker wellbeing and job attractiveness. In the following, we describe an ongoing multi-pronged effort to explore the potential of collaborative robots in the context of airplane engine repair and maintenance operations.

## I. INTRODUCTION

While it is often claimed that emerging robotic capabilities will help address labor shortages, the authors of this paper argue this is easier said than done. This abstract presents a transdisciplinary approach [1], [2] for developing robots on the work floor, with and for workers - in a specific use case for robot-assisted repair and maintenance.

Recent technological advancements have given rise to collaborative robots ("cobots") [3], which offer several advantages over traditional industrial robots, including safety, cost reductions, and flexibility. These benefits make cobots increasingly attractive to companies, leading to their steadily growing market share [4]. Despite their name, much work remains to be done before they can truly effectively collaborate with workers [5].

Physical Human-Robot Interaction (pHRI) [6], [7] provides a means to combine the cognitive capabilities of workers with the physical capabilities of robots, such as endurance, strength, or precision, to create a more efficient, effective and flexible workforce. In addition, pHRI has the potential to enhance worker safety, health, and productivity in various industrial applications including: reducing worker stress and workload [8], [9], managing muscle fatigue [10], [11], and improving ergonomics [12], [13]. While concerns around safety and ergonomics are crucial, one key aspect of robotic integration is ensuring that the collaborative work supports the psychological wellbeing of the workers, which has been given insufficient attention in HRI research [14].

In this paper, we briefly describe workplace-inspired lab research and prototyping with workers as an integrated effort towards transdisciplinary practices that may address these challenges. We specifically examine the aircraft engine repair and maintenance use case. See Fig. 1 for the overview of the proposed transdisciplinary approach in this use case.

## II. INVESTIGATION OF HUMAN-ROBOT ROLES

The introduction of cobots on the work floor has raised concerns about their impact on workers' psychological needs and job satisfaction. Studies within the social sciences show the (potential) impact of introducing AI and robots into the workplace. This includes impacts on work design [15], [16], job quality [17], work preferences [18], basic psychological needs [19], and meaningful work [20], [21], which in turn impact job satisfaction and psychological wellbeing of the workers. However, much of this work is either hypothetical or done in retrospect.

As cobots and human-robot interactions become more prevalent on the work floor, it is essential to investigate how they impact the psychological needs of workers. It is important to design and study effective robot capabilities that support the psychological wellbeing of workers in the context of robot-assisted physical work before introducing them into the workplace. As such, policymakers in the European Union are proposing a shift towards Industry 5.0 [22], which prioritizes a long-term sustainable future of work centered around workers and the planet, rather than just technology-driven smart factories that require minimal human intervention, as advocated for in Industry 4.0 [23]. A first step to approach achieving this vision is understanding the effects of the design of the collaborative system and worker-robot roles. While roles in pHRI have been investigated in the literature [24], [25], key fundamental insights into their meaningfulness and impact on psychological wellbeing are limited.

To address this gap, a transdisciplinary approach has been employed, involving iterative exchanges between the robotic lab environment and the real-world workplace, to ensure the relevance and applicability of the results (see Fig. 1). During initial visits to the KLM Engine Services workplace, maintenance tasks such as fan blade blending (which require human expertise but may still benefit from robotic assistance), were identified as ideal for investigating human-robot collaboration. This task was abstracted into a

This work was supported by the BrightSky project, funded by the R&D Mobiliteitsfonds from the Netherlands Enterprise Agency (RVO) and commissioned by the Ministry of Economic Affairs and Climate Policy.

[1]Department of Cognitive Robotics, Faculty of Mechanical Engineering, Delft University of Technology, Delft, The Netherlands (nicky.mol@tudelft.nl, l.peternel@tudelft.nl, j.m.prendergast@tudelft.nl, d.f.forster@tudelft.nl, d.a.abbink@tudelft.nl)

[2]Department of Sustainable Design Engineering, Faculty of Industrial Design Engineering, Delft University of Technology, Delft, The Netherlands (a.ianniello@tudelft.nl, o.siebinga@tudelft.nl, sara.muscolo@uniroma1.it)

[3]RoboHouse, Delft, The Netherlands (d.zatyagov@tudelft.nl, k.a.nachenius@tudelft.nl, e.s.verhoef@tudelft.nl)

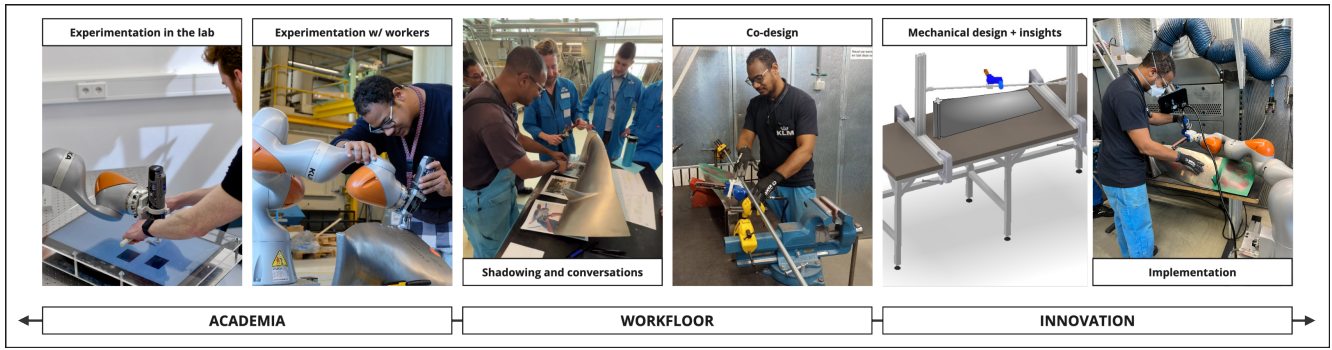

Fig. 1. Workflow of the proposed transdisciplinary research process in a jet engine repair and maintenance use case at KLM Engineering & Maintenance within the Brightsky project. The left two images show the research done at the academic level, which includes human-factor studies in lab environments. The middle two images show shadowing, observation, and co-design with workers on the workshop floor. The right two images show the innovation process taking place in both environments.

controlled lab setup, able to facilitate systematic user studies designed to investigate the hypothesis: *"How do different role distributions in a two-agent physical human-robot collaborative force-position task impact participants' perceptions of workload, system acceptance, and psychological wellbeing, specifically in terms of autonomy, competence, usefulness, and engagement?"*.

Initial findings from these studies reveal that varying role distributions influence participants' self-reported workload, system acceptance, and psychological wellbeing. Notably, different interaction strategies were observed among participants suggest the need for adaptive robotic support tailored to individual strategies. This evokes the development of personalized assistance, similar to modern automobiles, to enhance collaboration and worker satisfaction.

## III. PROTOTYPING WITH WORKERS

Building on the novel fundamental insights into human-robot roles, our approach then leverages the domain expertise of skilled workers early (and deeply) in the process. We consider this to be essential to the development of a successful technical and sustainable solution as early user involvement brings positive value for user and customer satisfaction that comes through better requirements [26]. After our first transdisciplinary visit, a cobot prototype was equipped with several HRI solutions (shared control, learning from demonstration) and demonstrated to workers on the workfloor. Feedback from worker interviews led to insights about further improvements such as changing hardware configuration and the setup layout for better ergonomics, software usability, and an alternative mechanical solution in lieu of robotization. Rapid prototyping techniques were immediately employed on-site.

Two potential solutions were developed: a mechanical system specifically tailored for the repair of a single type of fan blade, and a more flexible robotic setup with adjustable hardware and software configuration according to workers' individual way of doing the job (see Fig. 1). The designed mechanical system serves a dual purpose. First, it is a more economically viable alternative to a cobot for tasks that may not require the full complexity of a cobot system. Second, it is a way for workers without robotics expertise to more easily participate in co-design and communicate how the task should be done practically to the roboticists, which enables us to improve and better customize the cobot control strategies.

## IV. INTERACTION AND SPECULATIVE DESIGN

Meaningfully leveraging technology implies having a deep understanding of it and its implications when applied to the work context. In the proposed transdisciplinary research, this means digging into the differences in how engineering and design define robots and their capabilities, and parse HRI. This transdisciplinary knowledge can become a means to foster also speculative design processes [27] with the aim of understanding what would be relevant, appropriate, and desirable for the work organization and the workers. We employ speculative design to support the co-designing of visions shared across the work organization, leveraging technology and futures as critical conversations' enablers. Furthermore, it provides us with unconventional ways to bring innovative technologies to the workfloor even before their real application [28], making explicit the desires, expectations, fears, and concerns it might trigger [29].

## V. FUTURE WORK

The existing scope of the proposed transdisciplinary research approach is focused on specific use cases, such as the aircraft engine repair and maintenance showcased in this paper. We are currently actively involved in exploring this methodology also in two other use cases, i.e., nursing [30] and airport baggage handling. In the future, we will continuously evolve this transdisciplinary research paradigm and further expand the scope of our use cases. This compels us to include more disciplines and knowledge sources in our process beyond robotics, design, and hardware maintenance mentioned in this paper. Such disciplines and knowledge sources include but are not limited to ethics, psychology, economics, management, and organizational studies. Over the past years, we have managed to built up a strong consortium that unites researchers from these disciplines.

## REFERENCES

[1] M. v. d. Bijl-Brouwer, "Design, one piece of the puzzle: a conceptual and practical perspective on transdisciplinary design," *Proceedings of DRS*, 2022.

[2] C. Zaga, M. L. Lupetti, D. Forster, D. Murray-Rust, M. Prendergast, and D. Abbink, "First international workshop on worker-robot relationships: Exploring transdisciplinarity for the future of work with robots," in *Companion of the 2024 ACM/IEEE International Conference on Human-Robot Interaction*, ser. HRI '24. New York, NY, USA: Association for Computing Machinery, 2024, p. 1367–1369.

[3] J. E. Colgate, W. Wannasuphoprasit, and M. A. Peshkin, "Cobots: Robots for collaboration with human operators," in *Proceedings of the 1996 ASME international mechanical engineering congress and exposition*, 1996.

[4] I. F. of Robotics, "World Robotics Report: "All-Time High" with Half a Million Robots Installed in one Year," Oct 2022.

[5] J. E. Michaelis, A. Siebert-Evenstone, D. W. Shaffer, and B. Mutlu, "Collaborative or simply uncaged? understanding human-cobot interactions in automation," *Proceedings of the 2020 CHI Conference on Human Factors in Computing Systems*, 2020.

[6] S. Haddadin and E. Croft, "Physical human–robot interaction," *Springer handbook of robotics*, pp. 1835–1874, 2016.

[7] A. Ajoudani, A. M. Zanchettin, S. Ivaldi, A. Albu-Schäffer, K. Kosuge, and O. Khatib, "Progress and prospects of the human–robot collaboration," *Autonomous Robots*, vol. 42, pp. 957–975, 2018.

[8] C. Messeri, G. Masotti, A. M. Zanchettin, and P. Rocco, "Human-robot collaboration: Optimizing stress and productivity based on game theory," *IEEE Robotics and Automation Letters*, vol. 6, no. 4, pp. 8061–8068, 2021.

[9] V. De Simone, V. Di Pasquale, V. Giubileo, and S. Miranda, "Human-robot collaboration: an analysis of worker's performance," *Procedia Computer Science*, vol. 200, pp. 1540–1549, 2022.

[10] L. Peternel, C. Fang, N. Tsagarakis, and A. Ajoudani, "A selective muscle fatigue management approach to ergonomic human-robot co-manipulation," *Robotics and Computer-Integrated Manufacturing*, vol. 58, pp. 69–79, 2019.

[11] E. Merlo, E. Lamon, F. Fusaro, M. Lorenzini, A. Carfì, F. Mastrogiovanni, and A. Ajoudani, "Dynamic human-robot role allocation based on human ergonomics risk prediction and robot actions adaptation," in *2022 International Conference on Robotics and Automation (ICRA)*. IEEE, 2022, pp. 2825–2831.

[12] L. van der Spaa, M. Gienger, T. Bates, and J. Kober, "Predicting and optimizing ergonomics in physical human-robot cooperation tasks," in *2020 IEEE International Conference on Robotics and Automation (ICRA)*. IEEE, 2020, pp. 1799–1805.

[13] W. Kim, L. Peternel, M. Lorenzini, J. Babič, and A. Ajoudani, "A human-robot collaboration framework for improving ergonomics during dexterous operation of power tools," *Robotics and Computer-Integrated Manufacturing*, vol. 68, p. 102084, 2021.

[14] S. Fletcher, I. Eimontaite, P. Webb, and N. Lohse, ""We don't need ergonomics anymore, we need psychology!" – the human analysis needed for human-robot collaboration," in *Human Aspects of Advanced Manufacturing*. AHFE Open Acces, 2023.

[15] S. K. Parker and G. Grote, "Automation, algorithms, and beyond: Why work design matters more than ever in a digital world," *Applied Psychology*, vol. 71, no. 4, pp. 1171–1204, 2022.

[16] H. A. Berkers, S. Rispens, and P. M. Le Blanc, "The role of robotization in work design: a comparative case study among logistic warehouses," *The International Journal of Human Resource Management*, pp. 1–24, 2022.

[17] S. Baltrusch, F. Krause, A. de Vries, W. van Dijk, and M. de Looze, "What about the human in human robot collaboration? a literature review on hrc's effects on aspects of job quality," *Ergonomics*, vol. 65, no. 5, pp. 719–740, 2022.

[18] K. S. Welfare, M. R. Hallowell, J. A. Shah, and L. D. Riek, "Consider the human work experience when integrating robotics in the workplace," in *2019 14th ACM/IEEE international conference on human-robot interaction (HRI)*. IEEE, 2019, pp. 75–84.

[19] M. Gagné, S. K. Parker, M. A. Griffin, P. D. Dunlop, C. Knight, F. E. Klonek, and X. Parent-Rocheleau, "Understanding and shaping the future of work with self-determination theory," *Nature Reviews Psychology*, vol. 1, no. 7, pp. 378–392, 2022.

[20] J. Smids, S. Nyholm, and H. Berkers, "Robots in the workplace: a threat to—or opportunity for—meaningful work?" *Philosophy & Technology*, vol. 33, no. 3, pp. 503–522, 2020.

[21] S. Bankins and P. Formosa, "The ethical implications of artificial intelligence (ai) for meaningful work," *Journal of Business Ethics*, pp. 1–16, 2023.

[22] E. Commission, D.-G. for Research, Innovation, M. Breque, L. De Nul, and A. Petridis, *Industry 5.0 : towards a sustainable, human-centric and resilient European industry*. Publications Office, 2021.

[23] H. Lasi, P. Fettke, H.-G. Kemper, T. Feld, and M. Hoffmann, "Industry 4.0," *Business & information systems engineering*, vol. 6, pp. 239–242, 2014.

[24] A. Mörtl, M. Lawitzky, A. Kucukyilmaz, M. Sezgin, C. Basdogan, and S. Hirche, "The role of roles: Physical cooperation between humans and robots," *The International Journal of Robotics Research*, vol. 31, no. 13, pp. 1656–1674, 2012.

[25] L. Vianello, S. Ivaldi, A. Aubry, and L. Peternel, "The effects of role transitions and adaptation in human–cobot collaboration," *Journal of Intelligent Manufacturing*, pp. 1–15, 2023.

[26] S. Kujala, "User involvement: a review of the benefits and challenges," *Behaviour & information technology*, vol. 22, no. 1, pp. 1–16, 2003.

[27] J. H. Auger, *Why Robot? Speculative design, the domestication of technology and the considered future*. Royal College of Art (United Kingdom), 2012.

[28] M. L. Lupetti, C. Zaga, and N. Cila, "Designerly ways of knowing in hri: Broadening the scope of design-oriented hri through the concept of intermediate-level knowledge," in *Proceedings of the 2021 ACM/IEEE International Conference on Human-Robot Interaction*, 2021, pp. 389–398.

[29] P. Sengers, K. Williams, and V. Khovanskaya, "Speculation and the design of development," *Proceedings of the ACM on Human-Computer Interaction*, vol. 5, no. CSCW1, pp. 1–27, 2021.

[30] A. Arzberger, S. Menten, S. Balvert, T. Korteland, C. Zaga, E. Verhoef, D. Forster, and D. Abbink, "Barriers & enablers to transdisciplinarity in practice: Emerging learnings from a pilot project exploring the potential for robotic capabilities to improve nursing work," 10 2023.